\documentclass[letterpaper, 10 pt, conference]{ieeeconf}  

\IEEEoverridecommandlockouts                              

\usepackage{amsmath} 
\usepackage{amssymb}  

\usepackage{amsfonts}
\usepackage{algorithmic}
\usepackage{graphicx}
\usepackage{textcomp}
\usepackage{xcolor}
\usepackage[whole]{bxcjkjatype}
\usepackage[colorlinks]{hyperref}

\newcommand {\figref}[1] {Fig.~\ref{#1}}

\title{\LARGE \bf
Force Map: Learning to Predict Contact Force Distribution from Vision
}

\author{Ryo Hanai$^{1}$, Yukiyasu Domae$^{1}$, Ixchel G. Ramirez-Alpizar$^{1}$, Bruno Leme$^{1}$ and Tetsuya Ogata$^{2}$
\thanks{$^{1}$R. Hanai, Y. Domae, I.G.Ramirez-Alpizar, and B. Leme are with the National Institute of Advanced Industrial Science and Technology (AIST), Japan
        {\tt\small \{ryo.hanai, domae.yukiyasu, ixchel-remirezalpizar\}@aist.go.jp}}%
\thanks{$^{2}$T. Ogata is with the Graduate School of Fundamental Science and Engineering, Waseda University, Tokyo 169-8555, Japan and also with the National Institute of Advanced Industrial Science and Technology (AIST), Japan
        {\tt\small ogata@waseda.jp}}%
}

\begin{document}

\maketitle
\thispagestyle{empty}
\pagestyle{empty}

\begin{abstract}
When humans see a scene, they can roughly imagine the forces applied to objects based on their experience and use them to handle the objects properly.
This paper considers transferring this ``force-visualization'' ability to robots.
We hypothesize that a rough force distribution (named ``force map'') 
can be utilized for object manipulation strategies even if accurate force estimation is impossible. 
Based on this hypothesis, we propose a training method to predict the force map from vision. 
To investigate this hypothesis, we generated scenes where objects were stacked in bulk through simulation and trained a model to predict the contact force from a single image.
We further applied domain randomization to make the trained model function on real images.
The experimental results showed that the model trained using only synthetic images could predict approximate patterns representing the contact areas of the objects even for real images.
Then, we designed a simple algorithm to plan a lifting direction using the predicted force distribution.
We confirmed that using the predicted force distribution contributes to finding natural lifting directions for typical real-world scenes.
Furthermore, the evaluation through simulations showed that the disturbance caused to surrounding objects was reduced by 26 \% (translation displacement) and by 39 \% (angular displacement) for scenes where objects were overlapping.
\end{abstract}


\section{Introduction}
Based on current visual sensations and experiences, humans can make inferences about states that are difficult to measure directly and perform smart handling of objects.
For example, considering the state of force applied to objects piled up in bulk, they can pick an object such that the surrounding objects are not affected by the action or pack an object in a position and posture where the force does not affect the surrounding objects.
However, current vision-based robot object manipulation does not sufficiently consider such conditions and assumes the use of force and tactile senses after contact.
Such post-contact state understanding is insufficient for tasks such as handling a collapsible arrangement of objects or placing objects on non-fragile objects.

\begin{figure}[tbp]
\centerline{\includegraphics[width=\columnwidth]{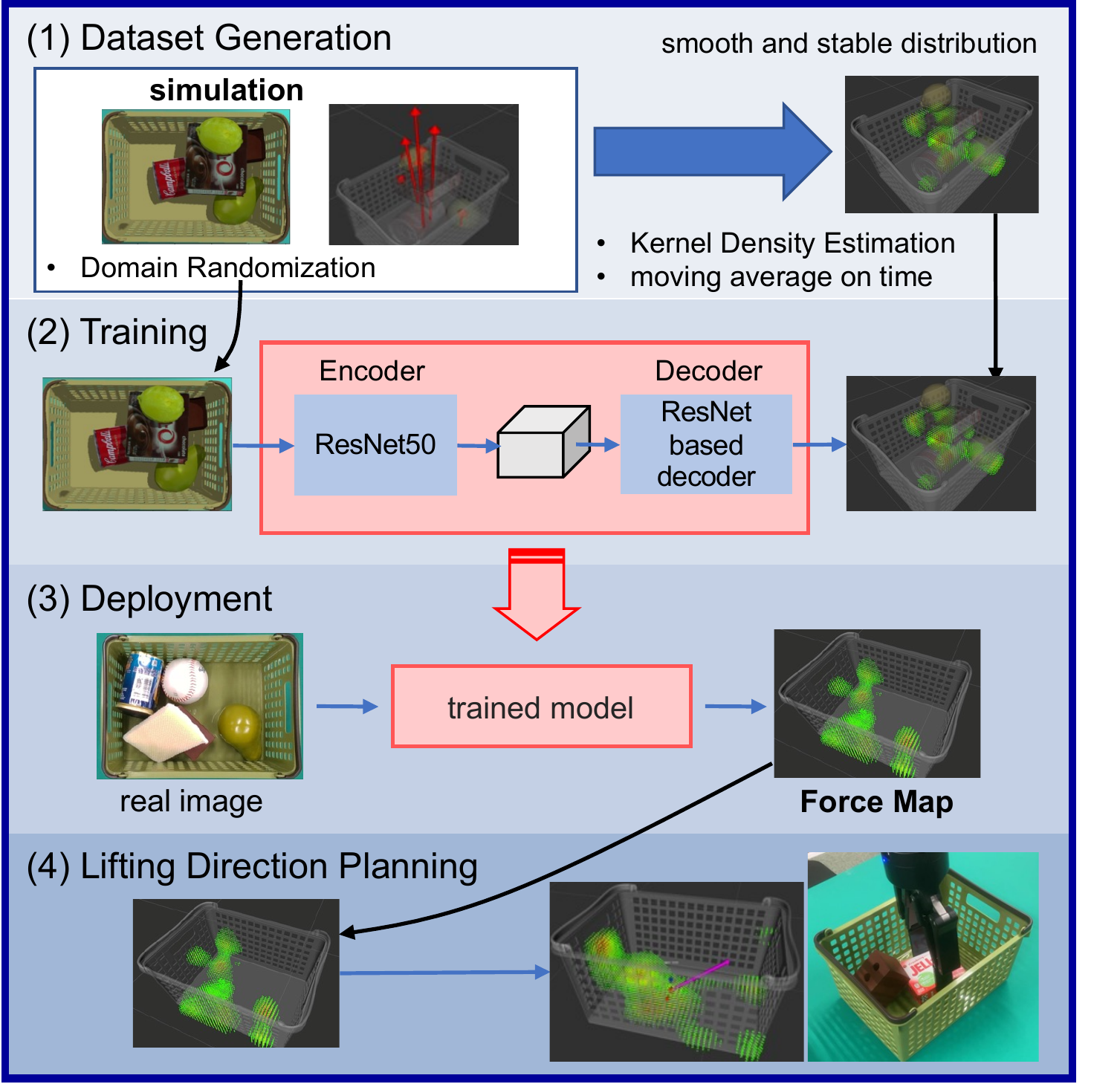}}
\caption{Overview of the proposed approach.
We train a model that predicts a rough distribution of contact forces from a single image. (1) Paired data of an image and corresponding contact forces are generated by simulation. The contact force data are converted to a smooth and stable distribution. 
(2) Training is performed using only the synthetic data.
(3) The trained model is used to predict a rough force distribution for real scenes.
The trained model generalizes to real-scene images because domain randomization was applied to the training images.
(4) Using the predicted force map, a lifting direction for picking a specified object is computed not to disturb the arrangement of surrounding objects.}
\label{fig:overview}
\end{figure}
To enable the inference of interactions among objects before contact, we propose a method to visually predict a {\it rough} distribution of forces applied to objects.
We use the word {\it rough} in this paper for two reasons.
First, the inference of force from vision is an ill-posed problem.
If we prepare an object that looks the same but has a different weight or hardness, 
our vision can be easily misled. This fact implies that it is impossible to always correctly infer contact forces from vision in general problem settings.
Apart from the intentional case, various packaged objects exist in our living spaces.
The weight of these objects changes as their content decreases with use, 
but their internal state cannot be visually determined from the outside.
Despite these limitations, we can extract helpful information for manipulation tasks 
if the application domain is appropriately restricted.
For example, in the case of a robot picking or filling a basket in a store, 
it is reasonable to assume that we can ignore the used products and the products are naturally stacked according to gravity.
Second, the contact force is very sensitive to contact differences; for example,
small-scale geometry and physical properties such as stiffness of the object surface.
Thus, we believe that the {\it rough} force distribution can be utilized for object manipulation strategies and validate the hypothesis experimentally.

Specifically, we consider the problem of predicting the contact forces acting on household items piled up from a single RGB image.
We propose a simulation-based method for training a model that predicts a rough distribution of contact forces from the visual input.
In the real world, it is difficult to measure forces. Direct measurement usually requires contact with an object.
Thus, the measurement is limited to the point a robot is touching, or a sensor is embedded.
This makes it difficult to infer the possible behaviors caused by the interaction of multiple objects.
However, simulations enable us to obtain sensory information that is not usually directly available in the real world.
In the use of simulation, the spotlight tends to be on how to bridge the gap between reality and simulation,
but we believe that the concept of acquiring experiences that are difficult to obtain in reality 
and do not require consideration of domain gaps is important\footnote{We call this methodology Experience Extension.}. 
Based on this concept, we aimed to enable inference for various real-world situations 
by training the model with simulation-generated synthetic data.

An overview of the proposed approach, which consists of four steps, is shown in \figref{fig:overview}.
In step 1, the simulation generates the contact force labels corresponding to images.
The contact force information obtained from the simulator was not directly used as the contact force labels; it was transformed into a spatially smooth and temporally stable distribution.
The details are described in Section \ref{subsec:contact_force_labels}.
In step 2, a cross-modal inference model was trained using the data generated in step 1.
In step 3, the trained model was applied to real images.
Domain randomization~\cite{Lee2021BeyondPT}\cite{Tobin2017DomainRF} was used to generalize the model trained with only synthetic data to real-scene images.
The actual forces between randomly stacked objects cannot be measured in the real world.
Therefore, we considered evaluating the usefulness of the estimated rough distributions of the contact force using a potential application. We present the results of planning the lifting direction for picking.

The first contribution of this study is that we proposed the concept of the “force map” as a rough distribution of contact forces inferred from vision. Second, we showed that a simple encoder-decoder model trained using only synthetic data can predict a reasonable three-dimensional (3D) contact force distribution even for real images of scenes where household objects are piled up. Lastly, we validated that the predicted contact force distribution has information helpful to plan a lifting function with less disturbance on the surrounding objects.

\section{Related Work}


Several previous works have explored force estimation from vision, but most of them focus on interactions between objects and a human: the contact force applied by the human hand to an object during object manipulation~\cite{Ehsani2020UseTF}\cite{8085141}\cite{Pham2015TowardsFS} or
the interaction force between objects and human bodies~\cite{Zhu2016InferringFA}\cite{Li2019Estimating3M}.
The estimation of the interaction force applied to objects assuming a situation 
where a robot gripper applies force to them~\cite{Shin2018SequentialIA}\cite{Hwang2017InferringIF} and
the estimation of the interaction force applied when a deformable object contacts another object at many contact points~\cite{Wang2022VisualHR}, have also been performed.
However, to the best of our knowledge, there has been no study on the estimation of contact force from the vision for a static scene with multiple objects interacting with each other.

This paper deals with the problem of estimating the interaction force between a group of objects stacked according to gravity.
Thus, this is a kind of scene-understanding problem.
The human cognitive ability to perform physical reasoning of objects from vision is called intrinsic physics, which has been studied in conjunction with the advancement of machine learning~\cite{Duan2022ASO}.
A typical task in physical reasoning for a static scene is stability prediction~\cite{Lerer2016LearningPI}\cite{Groth2018ShapeStacksLV}.
These works include determining the stability of a scene with objects of various shapes stacked on top of each other, 
predicting the point of stability as a heat-map, and predicting the outcome of physical interactions.
However, these works do not directly predict the interaction force.
Also, assumptions about physical properties, such as the uniformity of mass density, cannot be applied directly to a scene with various kinds of objects.

Visual information correlated with the force is necessary to estimate interaction force from vision.
Many of the above-mentioned studies use the movement~\cite{Ehsani2020UseTF}\cite{8085141}\cite{Pham2015TowardsFS}\cite{Li2019Estimating3M}
and/or deformation~\cite{Zhu2016InferringFA}\cite{Shin2018SequentialIA}\cite{Hwang2017InferringIF}\cite{Wang2022VisualHR}
as such visual cues.
However, these cues are unavailable for static scenes with no observable deformation.
Therefore, as explained in the introduction, we assume a statistical correlation between object type, appearance, and physical properties such as mass.
Such an assumption may not be sufficient for highly accurate estimation, but it can be utilized for rough inference.
In computer vision, inferring properties such as material and texture from appearance is common~\cite{Xiao2018UnifiedPP}. 
In robotics, the estimation of tactile properties~\cite{Takahashi2018DeepVL} and
stiffness~\cite{doi:10.1080/01691864.2022.2078669} from vision have been explored.
Our approach is end-to-end learning from appearance to physical interactions.
Although this approach does not explicitly estimate the properties such as mass, 
it is thought to correlate the appearance of objects to physical properties internally.


The essential part of the proposed method is generating
contact force information caused by contact between multiple
objects by simulation and using it for training. 
This sensory information is usually unavailable in reality. 
Such approaches are particularly effective in predicting information
that cannot be measured in the real world, such as the depth
of transparent objects~\cite{Sajjan2019ClearG3}. 
Such information is also utilized for the motion generation of robots~\cite{Lee2020LearningQL}. 
Although these approaches aim to obtain models
applicable to real-world input, it does not necessarily
mean that making the simulation close to reality is essential. 
Lee et al. trained the controller of a quadruped robot
using only rigid terrains and a small set of procedurally
generated terrains. However, the controller handled complex
real terrains, including deformable terrains~\cite{Lee2020LearningQL}. 
We also use a general-purpose simulator in which objects are modeled
as rigid bodies, and the geometries of objects are simplified but
we expect the trained model works for real scenes.




\section{Force Map}

\begin{figure*}[tbp]
\centerline{\includegraphics[width=0.8\textwidth]{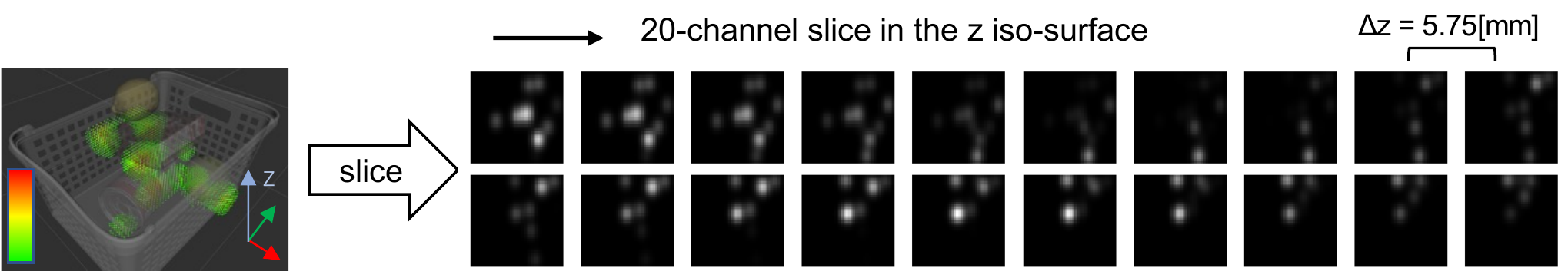}}
\caption{The force map is interpreted as a multi-channel image.}
\label{fig:slice}
\end{figure*}
\begin{figure}[tbp]
\centerline{\includegraphics[width=\columnwidth]{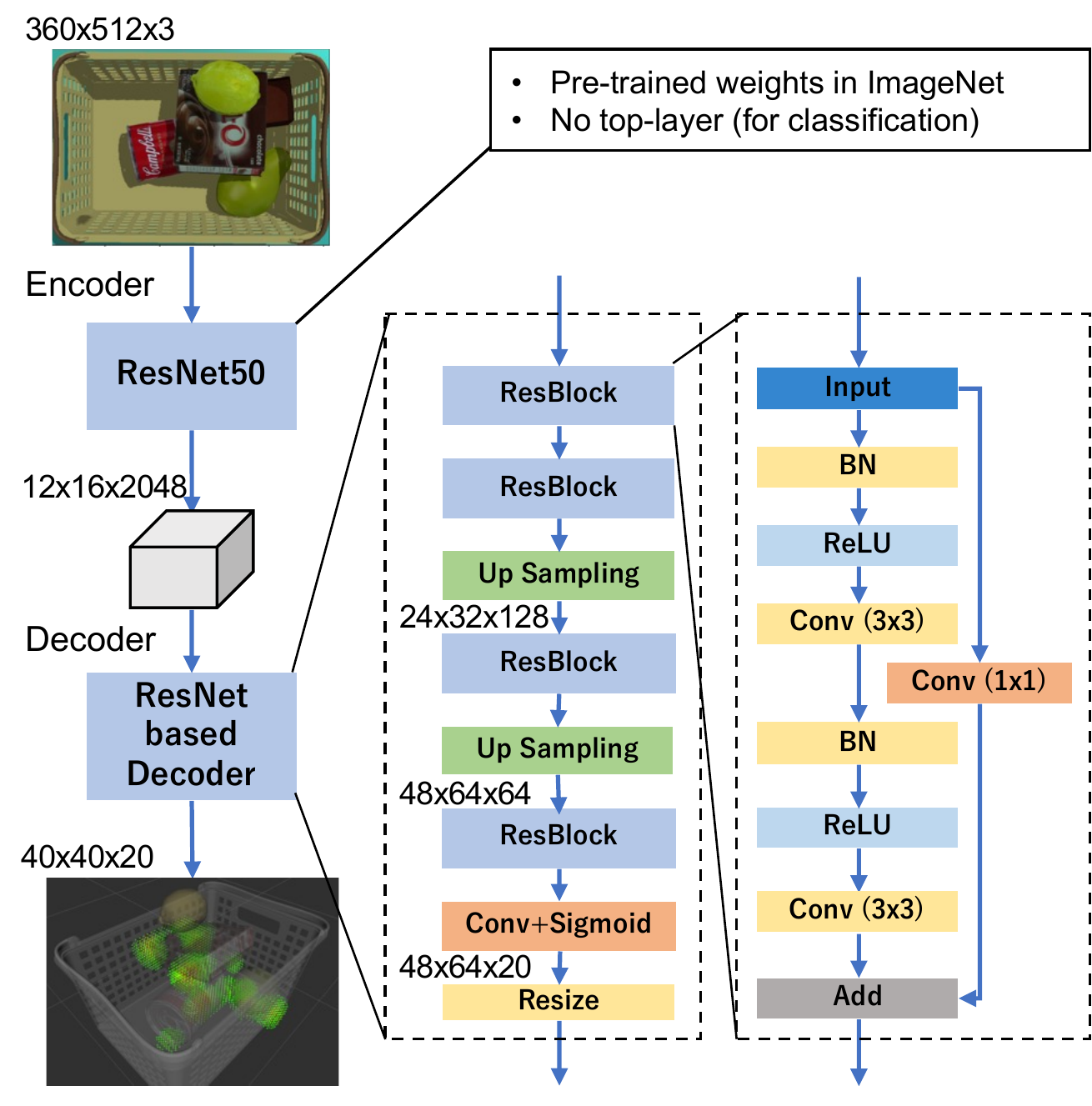}}
\caption{ResNet-based encoder-decoder model used for the prediction. 
This network translates an input RGB image into a multi-channel image representing a force map.}
\label{fig:network}
\end{figure}

\subsection{Basic Design}

There are various possible concrete representations of the force map.
For example, it could be an image-like two-dimensional representation viewed from the same viewpoint as the camera.
This study used low-resolution 3D voxels to represent the force map as shown in \figref{fig:overview}.
The purpose of the force map is not to represent forces with high accuracy and high resolution, but to represent the rough patterns of forces in the same way 
that humans are supposed to have when they manipulate.
For this reason, the resolution of the voxels was lower than that of the input RGB image.
The force map represents only the magnitude of the contact force observed at the coordinates of each grid point.
Estimating the 3D force distribution from a single RGB image is challenging.
Therefore, we did not distinguish the direction of the force but only represented its magnitude.

Next, we consider the problem of predicting a force map from a single RGB image. For this purpose, a general-purpose physics simulator for robotics was used to generate training data.
The training data were the paired data of an RGB image and the corresponding contact force distribution of the same scene.
Then, by using the generated paired synthetic data, we trained a model to predict the contact force distribution from the RGB image through supervised learning.

\subsection{Generation of the Contact Force Labels}\label{subsec:contact_force_labels}

We did not use the contact force information obtained from the simulator in its original form
 but converted it to a more spatially smooth and temporally stable distribution to generate the contact force labels.
This study used PyBullet(\cite{coumans2021}) as the physics simulator.
This paper assumes the use of PyBullet, but the same argument applies to many other general-purpose physical simulators for robots.

The expected contact force labels are the forces acting on objects in the {\it real world}.
However, it is difficult to measure these forces in a real scene.
We assumed that the simulator approximated them, but the output of the simulator differed from reality.
Generally, we did not expect to accurately predict the simulator’s output from the concept of the force map.
The contact information obtained from PyBullet is the position of contact points,
the magnitude of contact forces, and their direction vectors.
This information is spatially sparse.
The contact force varied depending on how the mesh of the object was structured and subtle differences in the contact state.
Such a behavior was not expected from the force map.
Furthermore, in this study, all objects were modeled as rigid bodies.
On the other hand, the surfaces of many daily objects have a certain degree of flexibility and are in contact with a certain area.
Therefore, their contact force distribution differs from the simulator's output, which is very sharp when the simulator calculates using the point-contact model for rigid bodies.
To reduce the effect that was clearly different from the expected behavior of the force map, spatial smoothing was performed.
General-purpose physics simulators usually aim to make the behavior of an object closer to reality over a certain time span.
The output of PyBullet sometimes fluctuates quickly, even when the object appears to be nearly stationary.
The contact force distribution is expected to be stable over time for static scenes. Otherwise, many different contact force distributions will appear for the same appearance.
To reduce this effect, smoothing along the time axis was also performed.

Specifically, we first applied the weighted Kernel Density Estimation (KDE)~\cite{Chen2017ATO} in 3D space.
Let the magnitude of the force at $n$ contact points $\bf{x_i}$ be $f_i$, and
the voxel spacing be h. The value of the force density at each voxel $\bf{x}$ is given as
\begin{eqnarray}
  \hat{f}(\bf{x}) &=& \frac{1}{nh^3}\sum^{n}_{i=1}f_iK(\frac{\bf{x}-\bf{x_i}}{\sigma}) \\
  K(\bf{x}) &=& \frac{1}{(2\pi)^{3/2}}e^{-\frac{\bf{\|x\|}^2}{2}},
\end{eqnarray}
where $K(\bf{x})$ is the Gaussian kernel and $\bf{\sigma}$ is the bandwidth parameter that controls the amount of smoothing. Since our objective is not the probability density estimation, $\hat{f}(\bf{x})$ is not normalized.
After the KDE, we took the moving average in time.

\subsection{Prediction of Force Map}

The model used for training is shown in \figref{fig:network}.
The target prediction task used a single RGB image as the input and a 3D distribution was the output.
To deal with this as an image-to-image translation problem,
we sliced the output distribution in the z-axis iso-surface and interpreted them
as a multi-channel image as shown in \figref{fig:slice}.
The training model was a residual neural network (ResNet)-based encoder-decoder model.
The encoder used the feature extraction part of ResNet50~\cite{He2015DeepRL} and excluded the top classification layer. The decoder consisted of residual blocks used by the
Residual U-Net~\cite{Zhang2017RoadEB} and a resize-layer to adjust the output image size.
The initial weight of the encoder was the pre-trained weight of ImageNet~\cite{Russakovsky2014ImageNetLS} and fine-tuning was performed.
Our final objective was to predict the distribution of contact forces in a real scene using the trained model.
Therefore, we applied data augmentation and domain randomization techniques so that the trained model generalized well to real scene images.

\subsection{Lifting Direction Planning}\label{subsec:lifting_direction_planning}

Because measuring the contact force applied to objects stacked in bulk in the real world is impossible, it is difficult to directly evaluate the accuracy of the predicted force map.
Instead, we validated the usefulness of the predicted force map by considering an application (described in this section).
We considered the problem of planning a lifting direction when a robot picks a specified object from a stack of objects not to disturb the arrangement of surrounding objects as much as possible.
In general, picking needs to consider other factors, such as grasp stability and collision avoidance in reaching; however, these factors are out of the scope of this paper.

\begin{figure}[tbp]
\centerline{\includegraphics[width=\columnwidth]{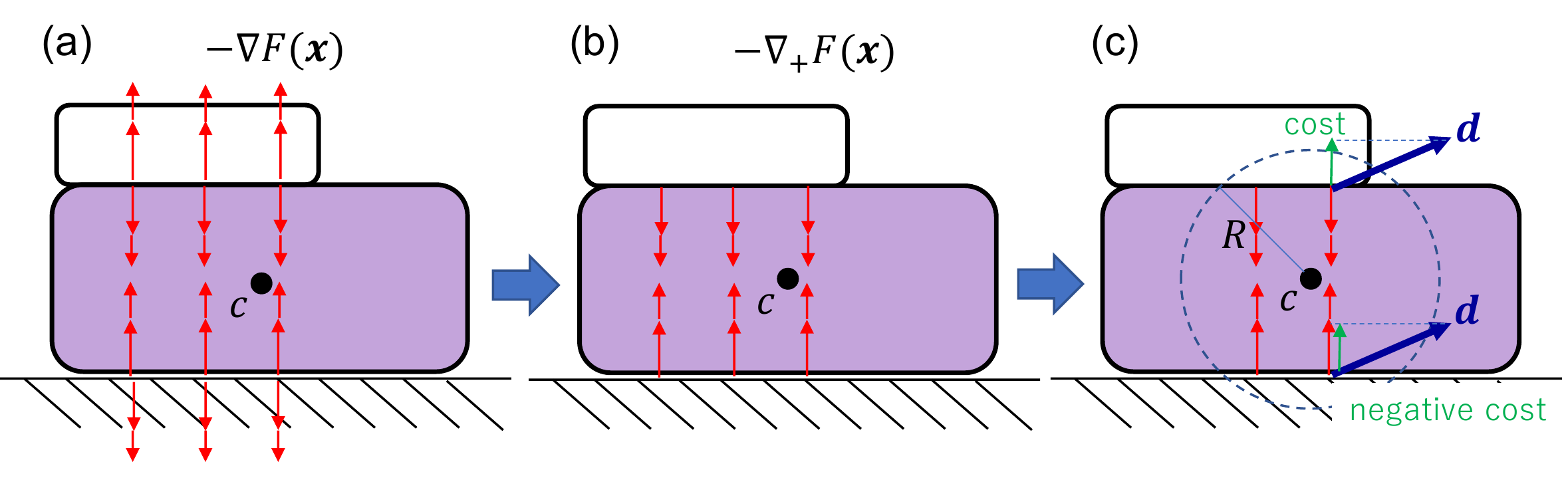}}
\caption{Lifting direction planning algorithm. (a) Force map takes large values on the contact surface.
(b) Force map gradient is restricted to those with an outward component from $\bf{c}$. Then, the negative value of the
restricted gradient approximates the force acting on the target object.
(c) Two costs are computed inside a sphere of radius $R$: cost for penetration and negative cost for repulsion.
}
\label{fig:picking_direction_method}
\end{figure}
We designed a heuristic algorithm to plan a lifting direction for a specified object using the predicted force map.
The force map itself had no information on the direction of the force; however,
it had a large value at the contact points on the target object's surface.
Consequently, when we looked at the object to be selected, 
the gradient of the force map around it was expected to approximate the normal forces applied to the object. 
However, this approach had a problem. 
Because we used only the information from the force map, we did not know the area of the object and its boundaries.
While combining the information from the force map with object recognition techniques was possible, in this study, we used the force map solely to compute the lifting direction.

\figref{fig:picking_direction_method} shows the image of the algorithm.
We assumed that the center ${\bf c}$ of the object $o$ to be selected was given.
The approximate size of the target object $R$ was also specified.
First, the gradient of the force map $F({\bf x})$ is restricted to those with an outward component from ${\bf c}$.
\begin{equation}
  \nabla_{+}F({\bf x}) = \left\{ \begin{array}{l}
          \nabla F({\bf x}) \;\;\; {\rm if} \; \nabla F({\bf x})\cdot ({\bf x}-{\bf c}) \geq 0 \\
          {\bf 0} \;\;\; {\rm otherwise}
         \end{array} \right.
\end{equation}
The negative value of $\nabla_{+}F({\bf x})$ approximates the force acting on the target object.
However, this still includes many forces generated at the boundaries between objects other than ${o}$.
Therefore, we limited the evaluation of the force map gradient to the interior of the sphere with center ${\bf c}$ and radius $R$.
Using $\nabla_{+}F({\bf x})$, the lifting direction vector ${\bf d}$ is computed to minimize the following
equation:
\begin{eqnarray}
  \text{min} && \int_{V({\bf x})}\text{LeakyReLU}(-\nabla_{+}F({\bf x})\cdot {\bf d}) dV \label{eq:lifting_direction_cost} \\
  \text{subject to} && \parallel {\bf d} \parallel = 1
\end{eqnarray}
Equation (\ref{eq:lifting_direction_cost}) considers two costs.
When an object is moved in the direction ${\bf d}$, the component against the gradient indicates that it pushes the contacting objects.
Because this should be avoided as much as possible, this component incurs a large cost. In addition, if there is no need to push the surrounding objects away, the object is expected to move away from the surface it is in contact with.
For example, if an object is placed on the floor, any movement in any direction except the floor direction component will not apply a force to the surrounding objects.
In this case, the object was expected to move away from the floor.
Therefore, we added a small negative cost to the component along the gradient when the object was moved in the direction of ${\bf d}$.
These two costs are combined in (\ref{eq:lifting_direction_cost}).
The negative region of Leaky Rectified Linear Unit (LeakyReLU) corresponds to the latter negative cost. 
The slope of the line in the negative region indicates the weight of the negative cost.

\section{Experiments}

\subsection{Dataset and Experimental Setup}

\begin{table}[tbp]
\caption{Parameter values used for the experiments}
\begin{center}
\begin{tabular}{c|c}
\hline
Property          & Value \\ \hline\hline
Force map grid spacing & 5.75 mm \\
KDE kernel size for contact force label smoothing & $\sigma$=12 mm \\
Target object radius & 5 cm \\
LeakyReLU slope for lifting direction planning & 0.1 \\ \hline
\end{tabular}
\label{tabl:parameters}
\end{center}
\end{table}

The dataset was created using 17 objects included 
in the YCB dataset~\cite{Calli_2015} in the simulator.
The masses of the objects were set to the values published in the YCB dataset.
The lateral friction coefficient was set to 0.3, and the rolling friction coefficient was set to 0.01 for all objects.
The dataset was created with the following procedure.
First, six randomly selected objects were placed in random positions and postures. The objects were not dropped from a high position
but were lowered slowly with the simulator's dynamics turned off.
When the object came into contact with the already placed objects or the basket,
the simulator's dynamics were turned on.
Data were recorded after waiting for object placement to stabilize.
Domain randomization was applied before recording.
The details of domain randomization are presented in the Appendix.
This process was performed 1,800 times, and 5,400 scenes 
with four to six objects were recorded.
These 5,400 scenes were divided into ratios of 75 \%, 12.5 \%, and 12.5 \% for
training, validation, and testing, respectively.
The parameters used for the following experiments are summarized in Table \ref{tabl:parameters}.

\subsection{Qualitative Evaluation of Predicted Force Map}

\begin{figure}[tbp]
\centerline{\includegraphics[width=0.8\columnwidth]{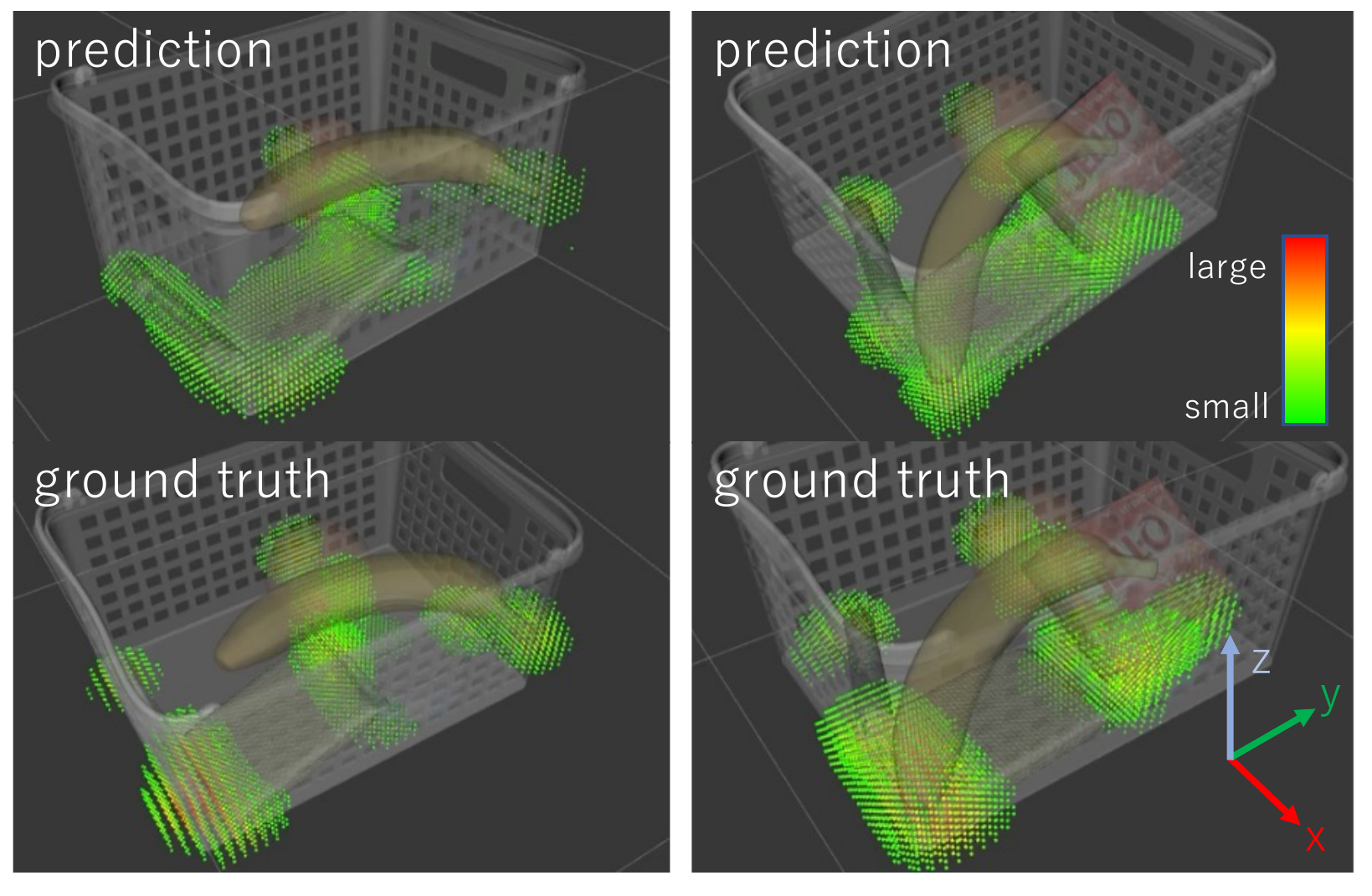}}
\caption{Predicted force maps for simulation scenes.
The color of each point indicates the magnitude of the force at that point. 
The color changes from green to red through yellow, indicating a larger force.
For clarity, points where the magnitude of the force was below a threshold were not drawn.}
\label{fig:prediction_sim}
\end{figure}
\figref{fig:prediction_sim} shows the results of the estimation of the test data generated by the simulation.
In these data, the objects are the same as the ones in the training data, but their arrangement is different from that in the training data. In \figref{fig:prediction_sim}, the predicted multi-channel image is visualized as a 3D point cloud.

\figref{fig:prediction_sim} shows that the overall force pattern was reasonably predicted.
Specifically, we can see that force is generated where the objects are in contact with the bottom of the basket, the walls of the basket, and with each other.
Although the estimation of a 3D distribution from a single image viewed from above is a difficult task, the position of contact between the wall and the object is predicted almost correctly in the z-direction.
However, the predicted distribution is generally blurred compared with the ground truth, and the peaks of the force are weaker.
Further, we observed a shadow-like false estimation at the bottom of the basket when there was a gap under an object (there is a gap under the banana in the left scene).
It is difficult to distinguish between cases where there is a contact on the bottom surface and cases where there is no contact based on slight changes in appearance.

\begin{figure}[tbp]
\centerline{\includegraphics[width=\columnwidth]{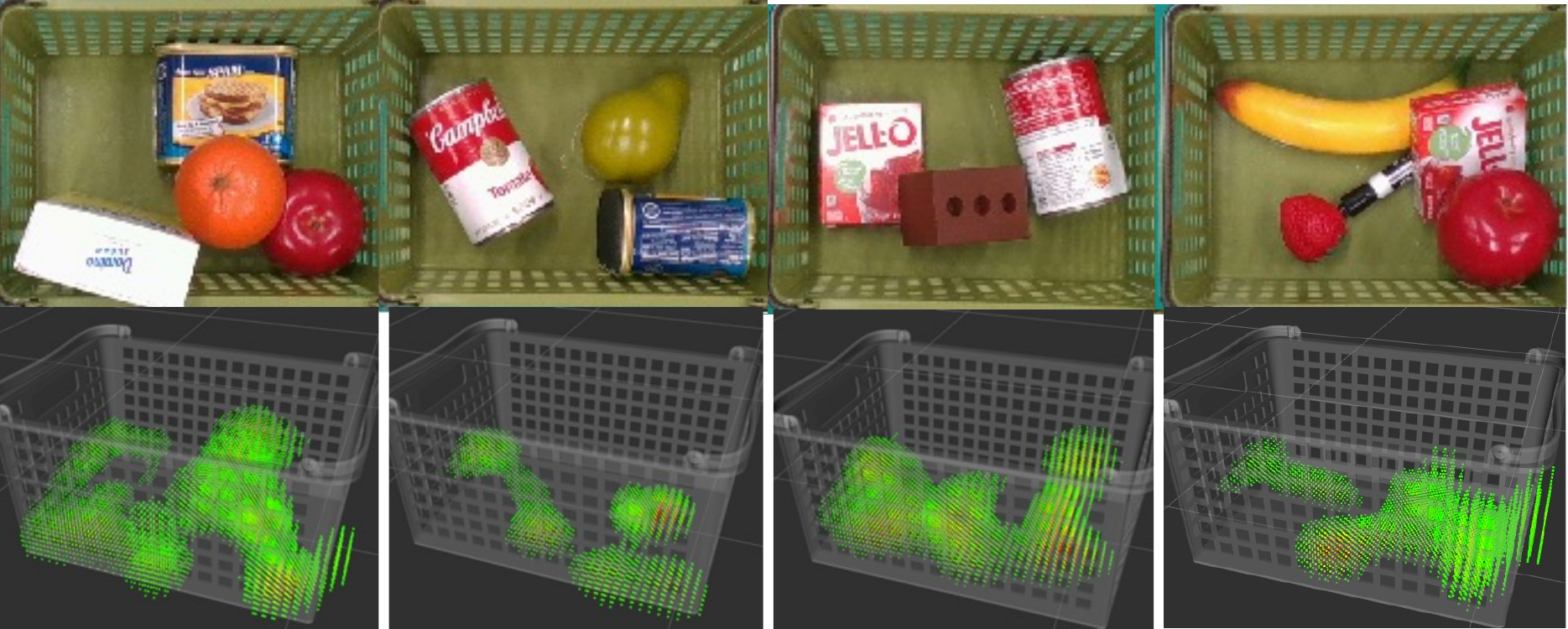}}
\caption{Predicted force maps for real scenes}
\label{fig:prediction_real}
\end{figure}
The estimation results for real-scene images are shown in \figref{fig:prediction_real}.
The 3D view shows only the estimated force map because the information on the objects in the basket is unknown in the real environment.
Although the training was performed using only simulation data, 
the trained model successfully predicted plausible force distribution
patterns for real scenes.
As in the case of the simulation data, the distribution was confirmed on the bottom surface, wall surfaces, and contact points between objects.
There was also some weak local noise, but the trained model worked for
a scene with a complex pile of objects.
Notably, the trained model did not work for real scenes without domain randomization.
\begin{figure}[tbp]
\centerline{\includegraphics[width=0.6\columnwidth]{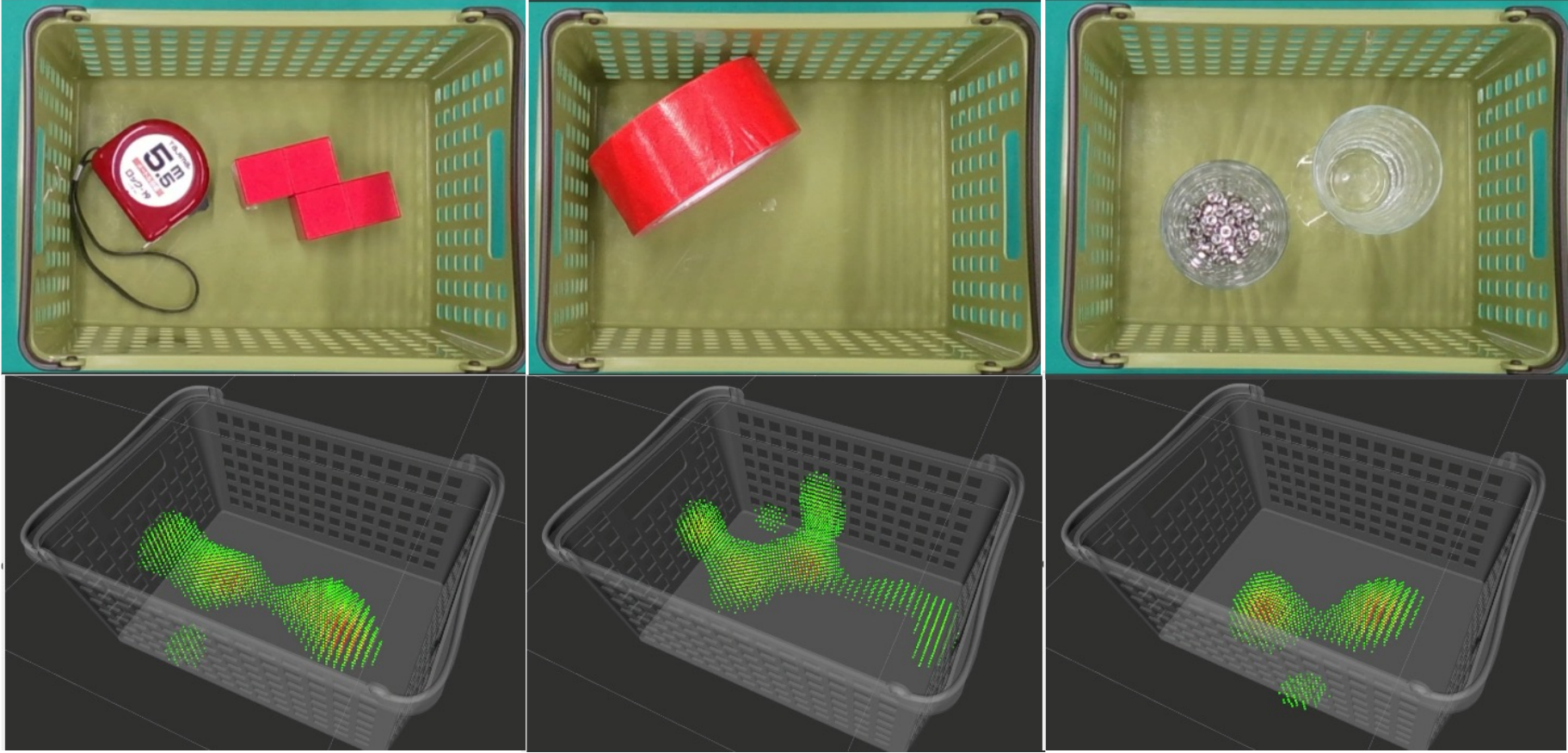}}
\caption{Predicted force maps for real scenes with novel objects}
\label{fig:novel_objects}
\end{figure}
In addition, estimation experiments were conducted on real-scene images with novel objects that were not included in the training data.
The results are shown in \figref{fig:novel_objects}.
The estimation of contact forces requires a certain amount of knowledge about the objects;
it is difficult to estimate the contact force from the knowledge of dozens of objects used for training.
However, we observed that the model captured the general pattern.
The model predicted the contact force where objects were placed,
as well as the contacts where the red packing tape was in contact with the wall surface (in the middle of \figref{fig:novel_objects}).
Even when a transparent glass not included in the training data was placed, the contact force was predicted at the bottom of the glass.

\subsection{Validation Using a Picking Application}

\begin{figure}[tbp]
\centerline{\includegraphics[width=\columnwidth]{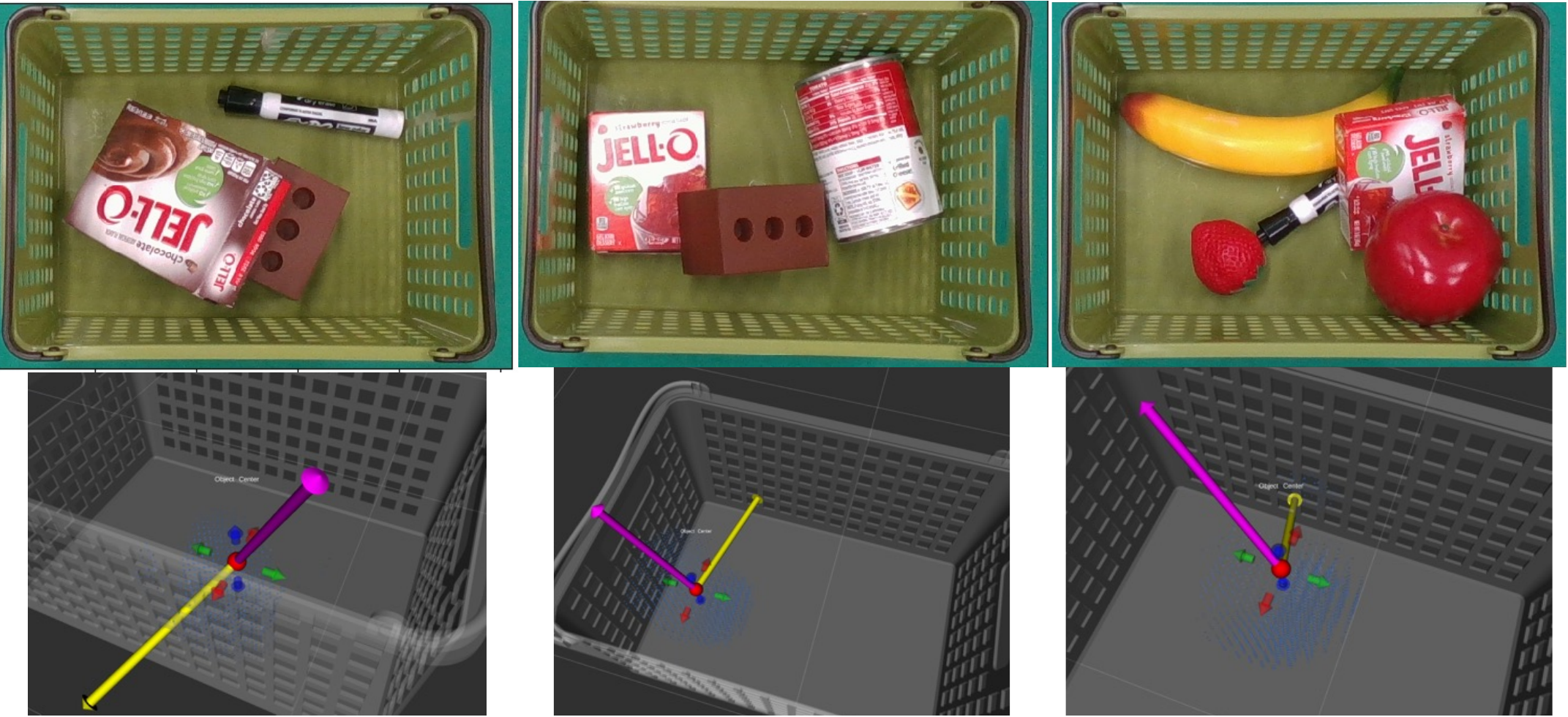}}
\caption{Planned lifting direction for real scenes}
\label{fig:picking_direction_results}
\end{figure}

\begin{figure*}[tbp]
\centerline{\includegraphics[width=0.66\textwidth]{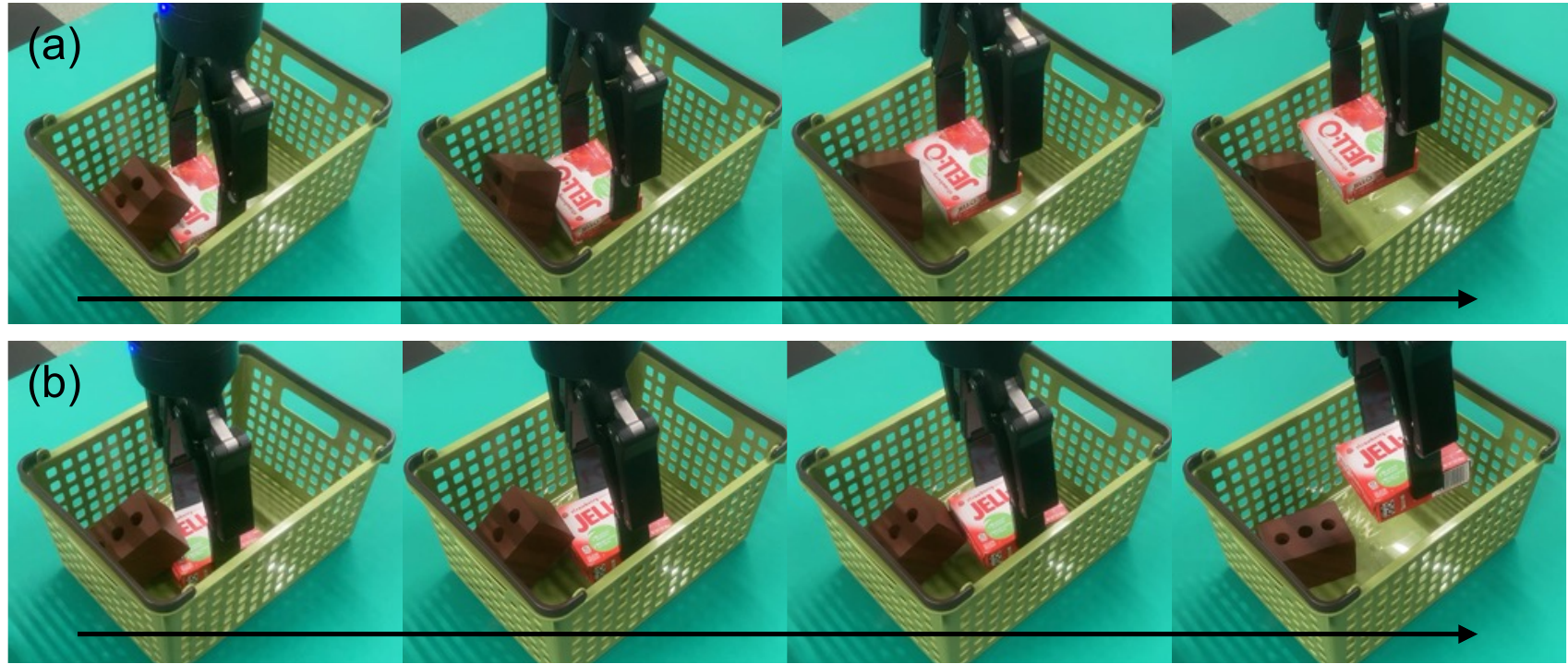}}
\caption{Lifting using a real robot. The robot picked the JELLO box. 
(a) When the robot lifted the box straight up, the brown rectangular solid moved significantly during the lifting and was placed vertically afterwards. (b) On the other hand, when the robot lifted the box in the calculated lifting direction, the movement of the brown rectangular solid was small throughout the lifting process, and the brown solid was placed in a pose close to the initial pose after the lifting.}
\label{fig:picking_real_robot}
\end{figure*}

The algorithm presented in \ref{subsec:lifting_direction_planning} was applied 
to some real scenes, as shown in \figref{fig:picking_direction_results}.
The lifting directions obtained by the algorithm are indicated by the magenta arrows.
The lifting target is the brown rectangle in the left scene,
the JELLO box in the middle scene, and the banana in the right scene.
The directions to move the target object
away from the surrounding objects were computed in all the scenes.
From the result in \figref{fig:picking_direction_results}, we can interpret that the predicted force maps for the real scenes contain useful information for picking.

We also performed lifting using a real robot.
The robot's trajectory was generated to move the object a certain distance in the calculated lifting direction and then lift it straight up.
The grasping position by the gripper was given independently because only the object's motion was planned from the force map.
As a baseline, we also performed an experiment in which the robot lifted the object straight up. As shown in \figref{fig:picking_real_robot}, the robot picked up the JELLO box below the brown rectangular solid.
When lifting the object straight up, the brown object moved significantly during the lifting; after the lifting, it was placed vertically, unlike its initial pose.
However, when the robot lifted the JELLO box in the calculated lifting direction, the movement of the brown object was small throughout the lifting process, and it was placed in a pose close to the initial pose after lifting.

\subsection{Quantitative Evaluation of Lifting Direction Planning}

We further investigated the quality of the planned lifting directions for
various configurations of objects in the simulation.
We used the displacement of objects from the initial position as an evaluation index. 
Because we were interested in the disturbance to surrounding objects,
we evaluated the displacement of objects other than the lifting target.
The displacement was measured as the maximum distance from the initial position during the lifting motion. The linear and angular distances were evaluated separately.
The proposed method first moved the target object by 5 cm in the planned direction and then lifted it straight up.
The baseline was the same as in the previous subsection and lifted the target straight up from the beginning.
Lifting was performed once for all objects in each of the 100 scenes,
with 4-6 objects randomly placed. The lifting direction was calculated 
using the center coordinates of the 3D model of the object.

\begin{figure}[tbp]
\centerline{\includegraphics[width=\columnwidth]{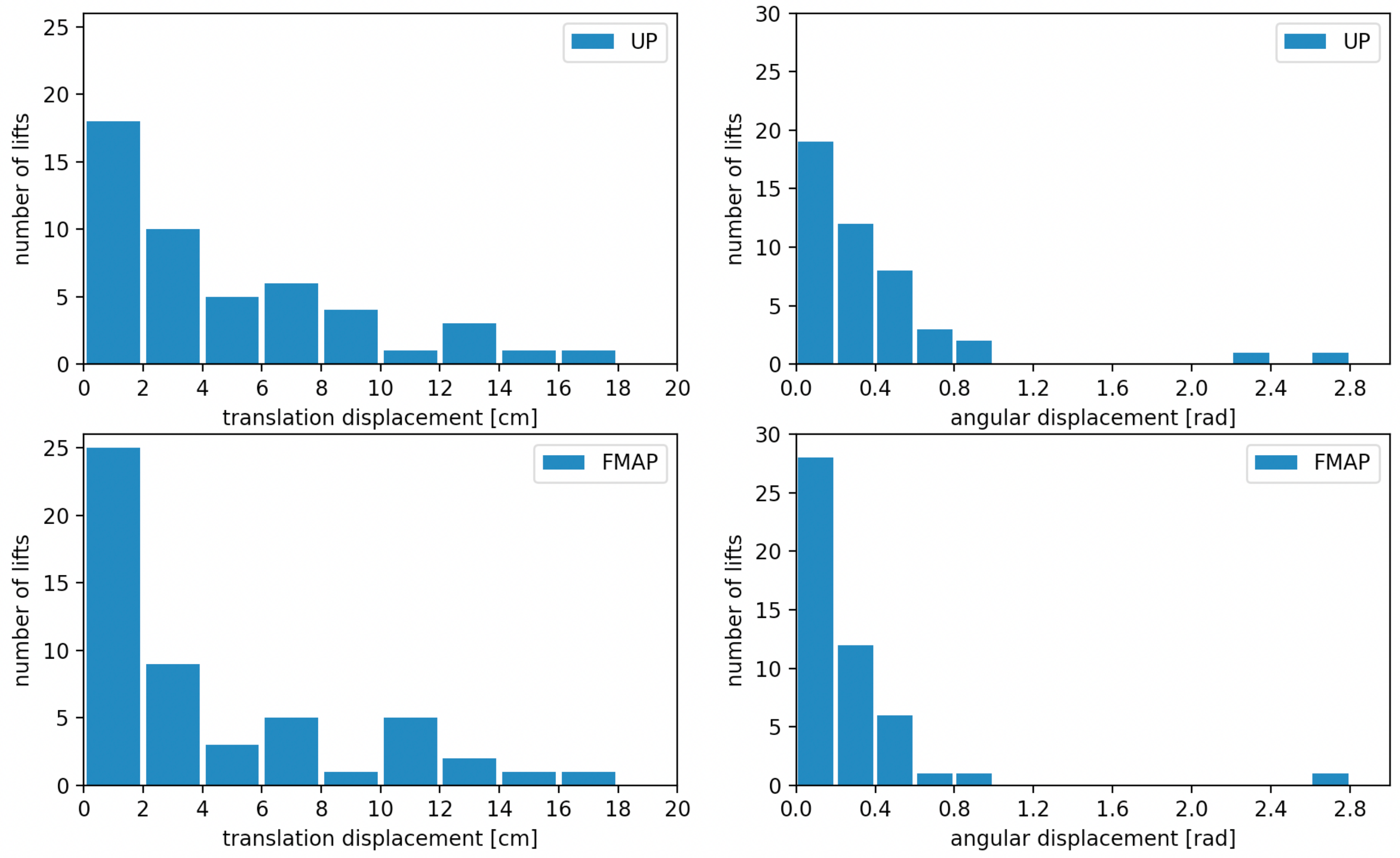}}
\caption{Distributions of the displacement when planned directions differ from upwards. FMAP was not effective only in special cases but reduced displacement overall.}
\label{fig:lifting_result_hist}
\end{figure}

Table \ref{tab1:lifting_direction_planning_result} presents the results.
FMAP indicates the proposed algorithm and UP indicates the baseline.
A total of 499 lifting trials were performed. The numbers in Table \ref{tab1:lifting_direction_planning_result} are the average values of the lifting trials.
The linear displacement showed a decrease of approximately 10 \% when using the force map.
On the other hand, the angular displacement worsened slightly.
In practice, the strategy for lifting straight lines was powerful. 
In particular, when the objects were aligned flat, trying to move the target objects horizontally　often increased the risk of interference with neighboring objects.
The second row of Table \ref{tab1:lifting_direction_planning_result} shows the trials in which the planning using the force map output significantly different directions from upwards.
We extracted the 52 lifting trials with an angle of 30 degrees or less with the bottom surface.
In this case, there was a significant decrease in both the linear and angular displacements. In other words, the force map effectively determines the direction to reduce the disturbance for scenes in which lifting in a different direction is effective.
In fact, the standard deviations of the displacements were quite large because
the distributions of the displacements were far from the normal distribution.
\figref{fig:lifting_result_hist} shows the distributions of the second row (the trials where the planned directions were different from upward).
It can be seen that FMAP was not effective only in special cases but reduced displacement overall.

\begin{figure}[tbp]
\centerline{\includegraphics[width=\columnwidth]{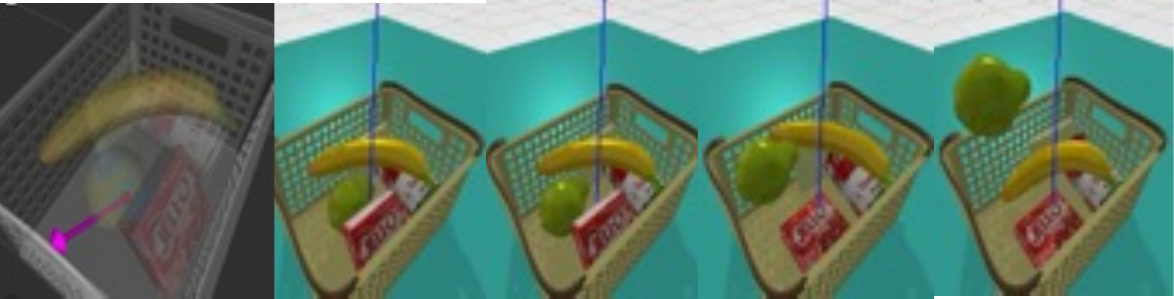}}
\caption{A typical successful case of lifting using the force map}
\label{fig:successful_case}
\end{figure}

Through the experiment, a typical pattern that benefits significantly from the force map method was observed.
In this case, the target was overlapped by other objects, and there was space to pull out the target horizontally or diagonally.
A typical successful case is shown in \figref{fig:successful_case}.
One of the limitations of the force map-based prediction is the presence of objects that are slightly distant. 
Theoretically, no contact force is generated between such objects and the target object. However, since the proposed method does not strictly determine contact, and a weak distribution appears between the objects and the target. This effect contributes to reducing the possibility of moving the target toward such close objects.
There is also a typical case in which the presented lifting direction planning method
does not work. The proposed method does not take object shape into account. 
In our experiments, we observed a case in which another object was on the edge of the banana. In this case, the method failed to consider the force applied by an object on the banana, and the banana was lifted straight up.

\begin{table}[tbp]
\caption{Results of lifting from random scenes through simulation}
\begin{center}
\begin{tabular}{c|c|c|c}
\hline
               & Method & Max linear       & Max angular \\
               &      & displacement(cm)$\downarrow$ & displacement(rad)$\downarrow$ \\ \hline\hline
All objects    & UP   & 3.40 $\pm$ 5.17 & 0.276 $\pm$ 0.552 \\
               & FMAP & 3.06 $\pm$ 4.55 & 0.300 $\pm$ 0.618 \\ \hline
Objects lifted in & UP   & 5.63 $\pm$ 5.55 & 0.710 $\pm$ 0.984 \\
different directions & FMAP & {\bf 4.15} $\pm$ 4.62 & {\bf 0.430} $\pm$ 0.769 \\ \hline
\end{tabular}
\label{tab1:lifting_direction_planning_result}
\end{center}
\end{table}

\section{Discussion}

The proposed force map prediction method has low accuracy 
for regions that are not visible due to occlusion.
Specifically, a false positive force distribution often appears when there is a gap under or behind an object.
To solve this problem, the use of multiple views is helpful.
Depending on the setup of the problem, it may be possible to obtain information
regarding the placement of the invisible area.
For example, in the case of a packing task, where the objects are packed 
in sequence, time-series information provides clues about the placement 
of the hidden area.

The closeness of the predicted force map to reality depends on the closeness of the simulation prediction to reality.
In this study, all objects were treated as rigid bodies. 
In recent years, simulation techniques for non-rigid objects, including deformable objects, have been improved and used for simulation of robotic tasks.
The use of such simulation techniques can improve the quality of the predicted
force map.
However, there are limits to making the simulation close to reality.
Using measurement data from a real environment is worth considering in this case.
We mentioned that measuring the contact forces acting on objects piled up in pieces
in the real environment is very difficult, if not impossible.
However, it is possible to measure the contact force with certain parts of the environment, such as the floor, using pressure sensors.
Thus, the combination of the simulation model and partially measured 
data is an approach worth exploring.


In the previous section, we used a force map for picking. 
Still, there are several other possible applications.
One of them is the packing mentioned in the introduction. 
Knowing the rough force acting on the objects makes planning how to pack possible,
so that the packed objects are not subject to a large force.
In addition, packing with little deformation may be achieved by combining the estimated stiffness of objects in the scene~\cite{doi:10.1080/01691864.2022.2078669}.

\section{CONCLUSIONS}

In this study, we hypothesized that predicting an approximate contact force distribution from the visual input is useful for object handling. To investigate this hypothesis, we trained a model to predict a 3D contact force distribution from a single image using only synthetic data generated by a general-purpose physics simulator.
The force labels of the training data were preprocessed to remove characteristics that clearly differed from those of the expected distributions.
The model used for training was a simple ResNet-based encoder-decoder model.
Domain randomization was employed to make the trained model applicable to real images.
We confirmed that the model predicted reasonable distributions for real images, 
even though it was trained using only synthetic data.
Furthermore, we designed a heuristic algorithm to plan the lifting direction using only the predicted distribution.
We also confirmed that the proposed algorithm could plan reasonable lifting directions for typical real scenes.
The algorithm reduced the linear displacement of the surrounding objects from 5.64 cm to 4.15 cm (by 26 \%) and the angular displacement from 0.710 rad to 0.430 rad (by 39 \%).
From these results, we conclude that the contact force distribution predicted by the proposed method contains useful information for object handling.
The process from the force map prediction to the lifting direction planning is lightweight and works in real time, taking an RGB image as input.


\section*{ACKNOWLEDGMENT}

We would like to thank Natsuki Yamanobe and Abdullah Mustafa of AIST for their useful discussions.
This work was supported by JST [Moonshot R\&D][Grant Number JPMJMS2031].




\section*{APPENDIX}

\subsection{Domain Randomization}

\begin{table}[tbp]
\caption{Domain randomization properties that are randomized and their ranges.}
\begin{center}
\begin{tabular}{c|c}
\hline
Property          & Range \\ \hline\hline
Lighting position & $(0, 0, 3) \pm (1, 1, 0.5)$ m \\
Shadow map intensity & $[0.2, 1.0]$ \\
Shadow map resolution & $[2048, 4096, 8192, 16384]$ \\
Shadow map world size & $[2, 10] \in \mathbb{Z}$ \\
Camera field of view & $21.25 \pm 1.0$ deg \\
Camera aspect ratio & $1.63294 \pm 0.1$ \\
Camera position & $(0,0,1.358)\pm(0.01,0.01,0.01)$ m \\
Camera target position & $(0,0,0.6)\pm(0.01,0.01,0.01)$ m \\
Camera up vector & $(-1,0,0)\pm(0.01,0.01,0.01)$ m \\
Object color & DefaultColor $\cdot [0.8\pm 0.2]$ RGB \\
Object specular color & $[(0,0,0), (1,1,1)]$ RGB \\ \hline
\end{tabular}
\label{tabl:domain_randomization}
\end{center}
\end{table}

\begin{figure}[tbp]
\centerline{\includegraphics[width=\columnwidth]{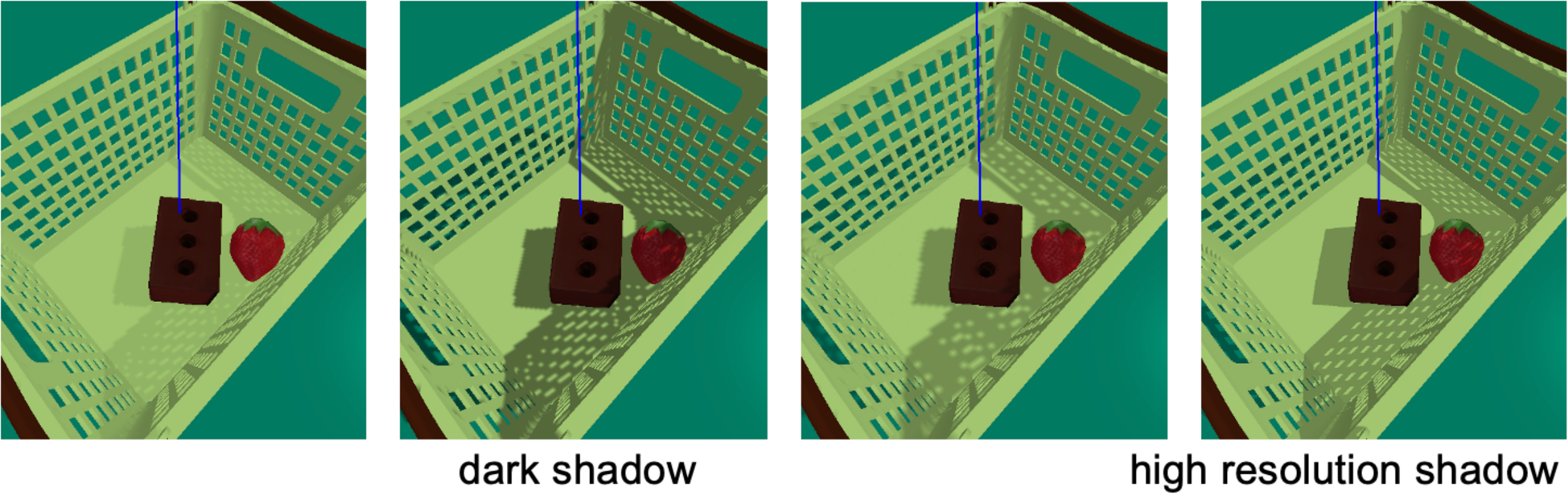}}
\caption{Domain randomization}
\label{fig:dr_shadow}
\end{figure}

The choice of visual domain randomization depends on the renderer.
PyBullet's default renderer was used.
Table \ref{tabl:domain_randomization} shows the randomized properties and their ranges.
Empirically, the influence of shadows was large, and false positive distributions were often observed in the shadowed areas.
Because we did not have many lighting options, we randomized the shading parameters to increase shadow variations.
The samples with shadow randomization are shown in \figref{fig:dr_shadow}.
Data augmentation was also applied to the domain randomized images during training.
The data augmentation added perturbations on brightness, contrast, hue in color, 
translation, rotation, zoom in geometry, and Gaussian noise.

\addtolength{\textheight}{-12cm}   
                                  
\end{document}